\documentclass{article} 
\usepackage{iclr2027_conference,times}

\usepackage{amsmath,amsfonts,bm}

\def\eqref#1{equation~\ref{#1}}

\def\1{\bm{1}}

\DeclareMathAlphabet{\mathsfit}{\encodingdefault}{\sfdefault}{m}{sl}
\SetMathAlphabet{\mathsfit}{bold}{\encodingdefault}{\sfdefault}{bx}{n}

\usepackage{hyperref}
\usepackage{url}
\usepackage{multirow}
\usepackage[utf8]{inputenc} 
\usepackage[T1]{fontenc}    
\usepackage{booktabs}       
\usepackage{amsfonts}       
\usepackage{nicefrac}       
\usepackage{microtype}      
\usepackage{xcolor}         
\usepackage{marvosym}

\usepackage{dsfont}

\usepackage{bm}
\usepackage{pifont}
\usepackage{algorithm}
\usepackage{algorithmic}
\usepackage{wrapfig}
\usepackage{colortbl}

\usepackage{amsmath}
\usepackage{amssymb}
\usepackage{mathtools}
\usepackage{amsthm}
\usepackage[most]{tcolorbox}
\usepackage{graphicx}
\usepackage{hyperref,xcolor}

\theoremstyle{definition}

\theoremstyle{remark}

\title{Formatting Instructions for ICLR 2027 \\ Conference Submissions}

\author{
Dayang Liang\thanks{Equal contribution.} \\
Xiamen University, Xiamen, China \\
\texttt{dyliang@stu.xmu.edu.cn}
\And
Lang Feng\footnotemark[1]~ \& Bo An \\
Nanyang Technological University, Singapore \\
\texttt{lang005@e.ntu.edu.sg}
\And
Yunlong Liu\thanks{Corresponding author.} \\
Xiamen University, Xiamen, China \\
\texttt{ylliu@xmu.edu.cn}
\AND
\textnormal{Project Page: \url{https://github.com/dy-liang/SAPO}}
}

\definecolor{midnightgreen}{rgb}{0.0, 0.29, 0.33}
\definecolor{deepgreen}{HTML}{055c29}
\definecolor{deeppurple}{HTML}{7030a0}
\definecolor{deepblue}{HTML}{171d91}
\definecolor{brown}{HTML}{843c0c}
\definecolor{shadered}{HTML}{ffe5e5}
\definecolor{shadegreen}{HTML}{e5f7ed}
\definecolor{msftBlack}{RGB}{0,0,0}
\definecolor{lightred}{RGB}{255,163,163}
\definecolor{deepred}{RGB}{153,0,0}

\definecolor{barblue}{RGB}{90,120,180}
\definecolor{barorange}{RGB}{225,124,5}

\newcommand{\methodname}{SAPO}

\definecolor{softblue}{RGB}{30, 90, 160}
\hypersetup{
  colorlinks=true,
  linkcolor=deepred,
  citecolor=softblue,
  urlcolor=magenta
}

\title{\textbf{SAPO}: Single-Rollout Autoregressive Policy Optimization for Agentic Reinforcement Learning}

\iclrfinalcopy 
\begin{document}

\maketitle

\begin{abstract}
Agentic reinforcement learning (RL) has emerged as an important post-training approach for enhancing the capabilities of Large Language Models (LLMs). However, existing methods face a trade-off  between policy performance and resource efficiency. Conventional Proximal Policy Optimization (PPO) implementations incur substantial memory overhead from a separate critic, whereas critic-free group-relative methods require multiple rollouts and face potential learning bottlenecks on long-horizon tasks. In this work, we propose \textbf{S}ingle-rollout \textbf{A}utoregressive \textbf{P}olicy \textbf{O}ptimization (\textbf{SAPO}), an efficient PPO-style framework that unifies policy optimization and value learning within a single causal language model, without . SAPO exploits the autoregressive structure of LLMs to sequentially generate action and value estimation at distinct causal boundaries with shared parameters, and then jointly optimizes the PPO objectives and an auxiliary on-policy SARSA objective with turn-level generalized advantage estimation, where the latter is designed to facilitate value learning. Extensive experiments on ALFWorld and WebShop with Qwen2.5-1.5B/7B and Qwen3-14B demonstrate that SAPO reduces peak GPU memory usage by \textbf{24.8\%} and per-iteration runtime by \textbf{23.1\%} over strong PPO baseline, while matching its task success rate. Our experiments also show that SAPO outperforms Group Relative Policy Optimization (GRPO) and recent cutting-edge variants in both task success and training stability.


\end{abstract}

\section{Introduction}
Recent advances in agentic reinforcement learning (RL) have transformed large language models from passive text generators into agents capable of long-horizon reasoning, tool use, web search, and code execution. Large-scale RL post-training has produced substantial improvements in mathematical reasoning and coding \citep{shao2024deepseekmath,deepseekai2025deepseekr1,hu2025openreasonerzero}, while interactive systems such as Search-R1 show that policies can learn to acquire external information through multi-turn interaction \citep{jin2025searchr1}. These successes suggest that learning directly from environmental feedback is becoming a central path toward more autonomous and general-purpose language agents.

Most agentic RL methods build on proximal policy optimization (PPO) or its critic-free, group-relative variants \citep{shao2024deepseekmath,yu2025dapo,li2026rethinking,hou2026sao}. PPO provides token-level or turn-level credit assignment through a learned value function and generalized advantage estimation (GAE), which possesses unique and even superior comprehensive learning capabilities, while typically requiring a separate critic model with substantial memory and computation costs \citep{ouyang2022training,stiennon2020learning,li2026turn}. Instead, GRPO removes this critic by normalizing rewards among multiple rollouts sampled for the same task \citep{shao2024deepseekmath}. However, homogeneous group rewards can yield vanishing advantages~\citep{yu2025dapo}, which can lead to performance bottlenecks and training instability. Its trajectory-level advantage also provides coarse temporal credit assignment and cannot reuse value estimates across states or policy iterations, unlike learned critics that can detect low-value prefixes~\citep{yuan2025vcppo,yue2025vapo,hu2025openreasonerzero}. Larger groups can stabilize estimates, but increase rollout and synchronization costs~\citep{choi2026poise,hou2026sao}.

Existing methods address different parts of this trade-off. For example,DAPO strengthens group-relative optimization through dynamic sampling and asymmetric clipping, but still relies on multiple rollouts and group statistics \citep{yu2025dapo}. ReMax~\citep{li2024remax} and RLOO~\citep{ahmadian2024back} replace the critic with greedy and leave-one-out baselines, respectively, while VinePPO~\citep{kazemnejad2024vineppo} estimates intermediate values using Monte Carlo continuations. These approaches either provide coarse temporal credit assignment or incur additional generation cost. Conversely, Open-Reasoner-Zero~\citep{hu2025openreasonerzero}, VC-PPO~\citep{yuan2025vcppo}, SAO~\citep{hou2026sao}, and VAPO~\citep{yue2025vapo} demonstrate the benefits of learned critics for long-horizon reasoning, but retain separate, often policy-scale value networks. Hydra-PPO~\citep{santacroce2023efficient} shares a frozen backbone but requires manually specified, role-specific adapters and heads, while POISE~\citep{choi2026poise} uses a lightweight probe on actor hidden states but still depends on cross-rollout estimation. Consequently, these leaves an open question: \textit{how can we retain the effective credit assignment of critic-based PPO while achieving the computational efficiency of group-based methods?}

We address this gap with Single-Rollout Autoregressive Policy Optimization (SAPO), which represents the state value and policy within a single causal language model, aiming to achieve high and stable learning performance at a lower computational cost. Specifically, SAPO exploits two natural boundaries in autoregressive generation, i.e., the prefix state before a response action provides the context for estimating $V(s)$, while the prefix extended by that action provides the context for estimating $Q(s,a)$. A shared two-logit value parameterization further extracts these estimates at their respective boundaries, where $V(s)$ provides the baseline for turn-level generalized advantage estimation and is trained with a value loss, whereas $Q(s,a)$ is trained with an auxiliary on-policy SARSA objective~\citep{rummery1994line} for potentially favorable value representations. The resulting advantages drive PPO’s clipped policy objective. Then, SAPO jointly optimizes these objectives using gradient masking, with scaled value learning and advantage normalization to improve training stability. In summary, SAPO turns the causal structure of language generation into an architectural principle for efficient value-based RL, retaining explicit temporal credit assignment within the resource profile of a single model and a single rollout.

We evaluate SAPO on ALFWorld and WebShop using Qwen2.5-1.5/7B and Qwen3-14B under matched optimization settings. Across model scales, SAPO matches powerful PPO in task success rate and learning variance while significantly outperforms GRPO. SAPO with only default terminal rewards is also comparable to state-of-the-art group-relative methods with additional dense reward designs. More importantly, SAPO reduces peak GPU memory usage by 15.2\%–35.6\% and per-iteration training time by 22.6\%–23.6\% compared with PPO. Therefore, SAPO provides a low-computation, single-rollout PPO-style training framework. Our contributions are threefold:

\begin{itemize}

   \item We introduce a single-rollout autoregressive policy optimization method that unifies policy optimization and value learning within a single causal language model.

   \item We systematically explore and incorporate mechanisms such as on-policy SARSA, scaled value learning, and dual logits to enhance the effectiveness of SAPO.

   \item Extensive experiments demonstrate that SAPO achieves a favorable combination of powerful performance, training stability, memory efficiency, and runtime efficiency.
\end{itemize}

\section{Related Work}

Agentic RL extends reinforcement learning with verifiable rewards from single-turn response optimization to multi-turn decision making, enabling LLM agents to learn from the consequences of actions taken in external environments. A prominent line of work builds on critic-free, group-relative optimization. GRPO~\citep{shao2024deepseekmath} replaces a learned value baseline with within-prompt outcome comparison, a formulation used by DeepSeek-R1~\citep{deepseekai2025deepseekr1} and refined by DAPO~\citep{yu2025dapo} with dynamic sampling and asymmetric clipping, GSPO~\citep{zheng2025group} with sequence-level importance ratios, and GVPO~\citep{zhang2025gvpo} through variance-aware gradient weighting. This paradigm has been extended to interactive settings: Search-R1~\citep{jin2025searchr1} and DeepResearcher~\citep{zheng2025deepresearcher} optimize iterative retrieval, ToRL~\citep{li2025torl} and WebSailor~\citep{li2025websailor} train tool use and web navigation, and RAGEN~\citep{wang2025ragen} characterizes optimization instabilities in multi-turn training. Long-horizon tasks have also motivated more structured credit mechanisms. ARChER~\citep{zhou2024archer} and HiPER~\citep{penghiper} decompose planning and execution across temporal scales, AgentGym-RL~\citep{xi2026agentgymrl} progressively increases the interaction horizon, and IGPO~\citep{wang2025igpo} derives dense intrinsic rewards from information gain during search.

Efficiency-oriented LLM RL has developed along critic-free advantage estimation, parameter sharing between policy and value functions, and systems-level execution. ReMax~\citep{li2024remax} and RLOO~\citep{ahmadian2024back} replace learned values with greedy or leave-one-out baselines, while GRPO~\citep{shao2024deepseekmath} normalizes rewards within each prompt group. Eliminating the value network reduces parameters and optimizer state, but these estimators require additional decoded trajectories and provide no value generalization across states. VinePPO~\citep{kazemnejad2024vineppo} instead estimates intermediate values through branched Monte Carlo continuations, trading critic training for rollout computation. Among parameter-sharing methods, Hydra-PPO~\citep{santacroce2023efficient} uses separate policy and value adapters on a frozen backbone; J-Hydra~\citep{santacroce2023efficient} shares the trainable adapter but risks interference between objectives. POISE~\citep{choi2026poise} predicts returns from actor hidden states but relies on cross-rollout estimation. SAO~\citep{hou2026sao} improves rollout utilization through asynchronous collection while retaining a separate critic. HybridFlow~\citep{sheng2025hybridflow} improves distributed execution through placement, resharding, and scheduling. SAPO instead represents the policy, state value, and action value in one causal model, avoiding group sampling and a separate critic while preserving temporal credit assignment.

\section{Preliminaries}
\paragraph{Multi-Turn Agentic RL.}
Multi-turn agentic RL trains an LLM policy through sequential interactions with an environment. We model each interaction episode as a finite-horizon Markov decision process (MDP), conditioned on a task instance $\boldsymbol{x} \sim p(X)$. At turn $t$, the policy samples a textual action $\boldsymbol{a}_t \sim \pi_\theta(\cdot \mid \boldsymbol{s}_t)$ from the current state $\boldsymbol{s}_t \in \mathcal{S}$. The environment then executes the action, returns a reward $r_t \in \mathbb{R}$, and transitions to $\boldsymbol{s}_{t+1}$. An episode with $T$ interaction turns produces the trajectory $\boldsymbol{\tau} = \bigl((\boldsymbol{s}_t, \boldsymbol{a}_t, r_t)\bigr)_{t=1}^{T}$. In our interaction format for tasks such as ALFWorld, each response comprises a reasoning segment enclosed in \texttt{<think>...</think>} and an executable command enclosed in \texttt{<action>...</action>}, only the command is passed to the environment. Feedback in these settings is often sparse and delayed, with an outcome reward $R(\boldsymbol{\tau}) \in [0, R_{max}]$ assigned upon episode completion.
\subsection{Group-relative Policy Optimization}

Recent agentic RL methods for LLMs often follow a group-relative policy optimization framework. For a task instance $\boldsymbol{x}$, the old policy $\pi_{\theta_{\rm old}}$ generates a group of $N$ complete rollouts, denoted by $\mathcal{G}_{x} = \{\boldsymbol{\tau}_i\}_{i=1}^{N}$. Each trajectory receives a scalar reward $R(\boldsymbol{\tau}_i)$ that measures task success or outcome quality. Whereas PPO typically uses a learned value critic to estimate advantages \citep{schulman2017proximal}, group-relative methods derive advantage estimates directly from the rewards of rollouts sampled for the same task:
\[
A(\boldsymbol{\tau}_i)
=
\left[
\mathtt{GroupNormalization}
\left(\{R(\boldsymbol{\tau}_j)\}_{j=1}^{N}\right)
\right]_i.
\]
In GRPO \citep{shao2024deepseekmath}, this estimate is obtained by subtracting the group mean from each trajectory reward and dividing by the group standard deviation. By using within-group reward statistics as the baseline, GRPO eliminates the need for a separate value critic, reducing the associated memory and computational costs.

\subsection{Proximal Policy Optimization}
Proximal Policy Optimization (PPO) is an actor--critic reinforcement learning algorithm widely used for LLM post-training \citep{schulman2017proximal}. Given prompts sampled from a dataset, the policy generates response trajectories that are evaluated by task-specific environments or learned reward models. The policy is optimized using a clipped surrogate objective:
\[
\mathcal{L}_{\mathrm{PPO}}
=
\mathbb{E}_t\left[
\min\left(
\rho_t \hat{A}_t,\,
\operatorname{clip}(\rho_t,1-\epsilon,1+\epsilon)\hat{A}_t
\right)
\right],
\]
where
$\rho_t = \pi_\theta(a_t \mid s_t) /
\pi_{\theta_{\mathrm{old}}}(a_t \mid s_t)$
is the likelihood ratio between the current and old policies,
$\hat{A}_t$ is the estimated advantage, and $\epsilon$ controls the clipping range. Clipping limits the incentive for large policy updates, helping stabilize optimization. A learned value critic $V_\phi(s_t)$ estimates expected future returns and supports advantage estimation, commonly through generalized advantage estimation. These estimates provide learning signals for earlier decisions when rewards are delayed. In token-level LLM optimization, states represent token prefixes or interaction histories, and actions correspond to generated tokens. Although the critic supports credit assignment across long trajectories, maintaining and training it introduces additional memory and computational costs.

\section{Method}
\begin{figure}[hb]
   \centering
   \vspace{-5pt}
   \includegraphics[width=\linewidth]{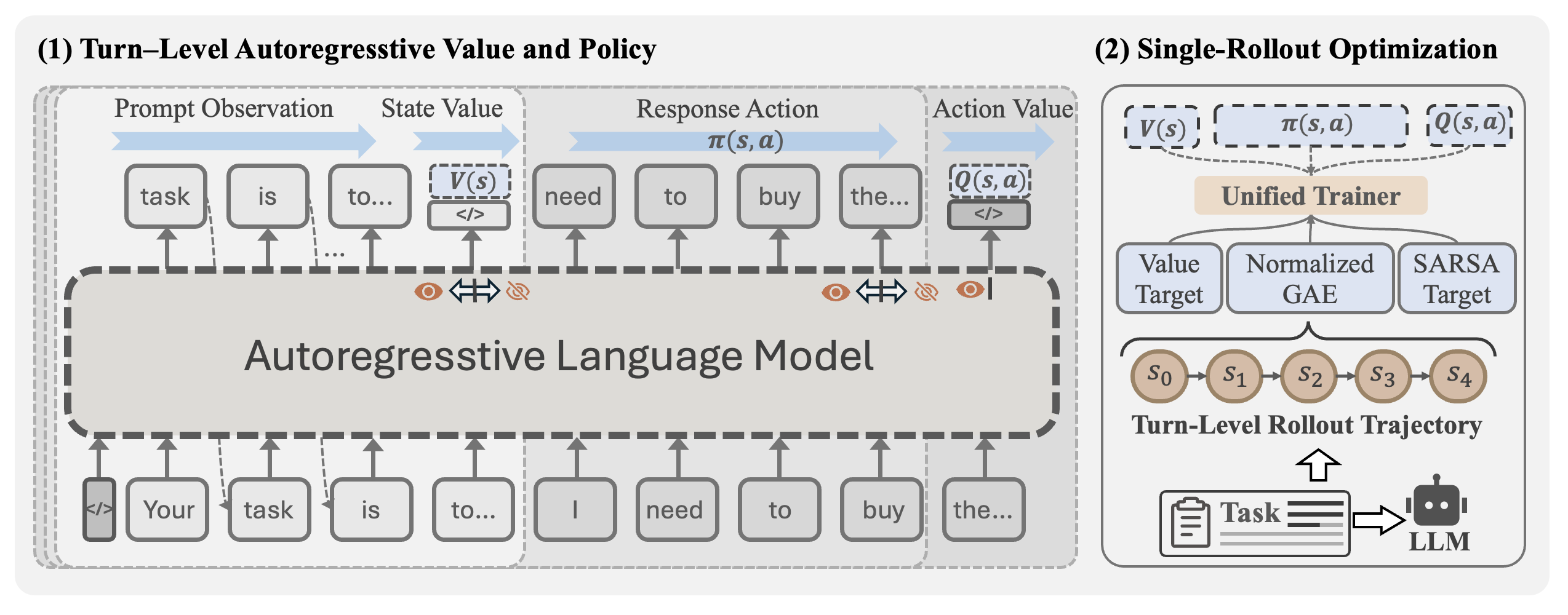}
   \vspace{-12pt}
   \caption{\textbf{Overview of SAPO.} \textbf{Left}: Value and policy autoregressive processes. A shared autoregressive language model generates actions and reads $V$ and $Q$ from logits at the pre- and post-action boundaries. \textbf{Right}: Loss structure and training pipeline. One rollout per task supplies turn-level value targets and advantages, which are broadcast to response tokens for policy optimization.}
   \label{fig:fw}
   \vspace{-10pt}
\end{figure}

We propose Single-rollout Autoregressive Policy Optimization (SAPO), an efficient PPO-style framework that unifies policy
optimization and value learning within a single causal language model. \methodname{} represents the policy, state value, and action value within one causal language model (Figure~\ref{fig:fw}). It reads values at the boundaries before and after each generated response, constructs turn-level advantages and value targets from one rollout per task, and jointly optimizes turn-level value objectives and token-level policy. Algorithm~\ref{alg:sapo} in Appendix~\ref{app:sapo-algorithm} summarizes the training procedure.

\subsection{Autoregressive Policy and Value Representation}
\label{sec:autoregressive-ac}

\paragraph{Causal Conditioning at Action Boundaries.}
At turn $t$, let $\boldsymbol{c}^{s}_t$ represent the current interaction history and $\boldsymbol{a}_t=(a_{t,1},\ldots,a_{t,M_t})$ the generated response. The state value conditions on $\boldsymbol{c}^{s}_t$, while the action value additionally conditions on the complete response, using $\boldsymbol{c}^{sa}_t=[\boldsymbol{c}^{s}_t;\boldsymbol{a}_t]$. We read these values at the final state-context position and the final response position, respectively. Causal masking prevents the state-value readout from accessing the current action and the action-value readout from accessing the subsequent reward or observation.

\paragraph{Shared Scalar Readout.}
Let $\boldsymbol{z}_\theta(\boldsymbol{c})\in\mathbb{R}^{|\mathcal{W}|}$ denote the next-token logits after context $\boldsymbol{c}$. We reserve two vocabulary entries, $w^+$ and $w^-$, and use their logit difference as a scalar value readout:
\begin{equation}
    v_\theta(\boldsymbol{c})
    =\frac{z_{\theta,w^+}(\boldsymbol{c})-z_{\theta,w^-}(\boldsymbol{c})}{\tau_v},~ v_\theta(\boldsymbol{c})\in\mathbb{R},
    \label{eq:value-readout}
\end{equation}
where $\tau_v>0$ controls the readout scale. The readout is unconstrained: we apply neither output clipping nor a probability mapping. Given a return-scale factor $R_{\max}>0$, the two boundaries yield
\begin{equation}
V_\theta(\boldsymbol{s}_t)=R_{\max}v_\theta(\boldsymbol{c}^{s}_t),~
    Q_\theta(\boldsymbol{s}_t,\boldsymbol{a}_t)=R_{\max}v_\theta(\boldsymbol{c}^{sa}_t).
    \label{eq:vq-readout}
\end{equation}

\paragraph{Value-Target Scaling.}
We train the readout against value targets divided by $R_{\max}$ (Section~\ref{sec:joint-objective}), while using the rescaled $V$ and $Q$ in Equation~\ref{eq:vq-readout} for temporal-difference and advantage computation. This separates the numerical scale of value regression from the reward scale used for policy learning. Empirically, with outcome rewards scaled to $10$, regressing scaled targets improves performance over directly fitting unscaled targets. We hypothesize that requiring large logit differences for value prediction perturbs the pretrained model's logit structure and increases interference with policy learning. Target scaling encourages smaller readouts without imposing a hard output range.

\paragraph{Action Distribution and Shared Evaluation.}
The reserved entries serve only as readouts and are never sampled or appended to the context. We exclude them from the action vocabulary, $\mathcal{W}_{\mathrm{act}}=\mathcal{W}\setminus\{w^+,w^-\}$, and define
\begin{equation}
    \pi_\theta(a\mid\boldsymbol{c})
    =\frac{\exp z_{\theta,a}(\boldsymbol{c})}
    {\sum_{w\in\mathcal{W}_{\mathrm{act}}}\exp z_{\theta,w}(\boldsymbol{c})},~a\in\mathcal{W}_{\mathrm{act}}.
    \label{eq:restricted-policy}
\end{equation}
We use this support for rollout sampling, policy likelihoods, and regularization. The actor evaluation on each sampled state--response sequence supplies both token log probabilities and the two value readouts. During optimization, these predictions likewise share one forward pass, eliminating the separate critic's forward/backward computation and training state. Policy and value losses still interact through the shared parameters.

\subsection{Turn-Level Advantages and Value Targets}
\label{sec:single-rollout-gae}

For each task, we sample one trajectory $\boldsymbol{\tau}=((\boldsymbol{s}_t,\boldsymbol{a}_t,r_t,d_t))_{t=1}^{T}$ with the rollout policy $\pi_{\theta_{\mathrm{old}}}$, where $d_t$ indicates episode termination. All targets use frozen rollout-time predictions, expressed on the original reward scale through Equation~\ref{eq:vq-readout}, and are held fixed during optimization.

\paragraph{Turn-Level Generalized Advantage Estimation.}
With $V_t^{\mathrm{old}}=V_{\theta_{\mathrm{old}}}(\boldsymbol{s}_t)$, we apply GAE along the interaction turns:
\begin{align}
    \delta_t&=r_t+\gamma(1-d_t)V_{t+1}^{\mathrm{old}}-V_t^{\mathrm{old}},
    \label{eq:td-residual}\\
    A_t^{\mathrm{GAE}}&=\delta_t+\gamma\lambda(1-d_t)A_{t+1}^{\mathrm{GAE}},
    \label{eq:trajectory-gae}
\end{align}
with $A_{T+1}^{\mathrm{GAE}}=0$. The task horizon defines episode termination in our finite-horizon setting, so the final bootstrap value is zero. This recursion propagates delayed rewards across turns, with $\lambda$ controlling the bias--variance trade-off.

\paragraph{State- and Action-Value Targets.}
We supervise $V$ with a $\lambda$-return and $Q$ with an on-policy SARSA target:
\begin{equation}
    y_t^V=V_t^{\mathrm{old}}+A_t^{\mathrm{GAE}},~
    y_t^Q=r_t+\gamma(1-d_t)Q_{t+1}^{\mathrm{old}},
    \label{eq:vq-targets}
\end{equation}
where $Q_{t+1}^{\mathrm{old}}=Q_{\theta_{\mathrm{old}}}(\boldsymbol{s}_{t+1},\boldsymbol{a}_{t+1})$ uses the next action from the same rollout; the bootstrap term is zero at termination. The state value supplies the baseline for GAE, while $Q$ provides auxiliary supervision after each response. This on-policy SARSA update adds negligible overhead, and its purpose is to strengthen state-value learning. Because $V$ and $Q$ use the same two reserved logit coordinates, gradients from post-action $Q$ regression shape the shared value representation used to produce the pre-action $V$ estimate. Table~\ref{ablation} empirically supports this coupling, as removing $\mathcal L_Q$ consistently degrades performance. Notably, action value errors do not directly enter the advantage estimate because $Q$ is not involved in the GAE.

\paragraph{Advantage Normalization.}
For environments with an invalid-action indicator $m_t^{\mathrm{inv}}$, we retain the adjustment $\widetilde A_t=A_t^{\mathrm{GAE}}-c_{\mathrm{inv}}m_t^{\mathrm{inv}}$; otherwise $c_{\mathrm{inv}}=0$. We compute the mean $\mu_{\mathcal B}$ and standard deviation $\sigma_{\mathcal B}$ of $\widetilde A_t$ over all valid turns in the rollout batch, and set$    \widehat A_t=\frac{\widetilde A_t-\mu_{\mathcal B}}{\sigma_{\mathcal B}+\epsilon_{\mathrm{adv}}} $. Statistics are computed before broadcasting advantages to response tokens, so each turn contributes once regardless of response length. This standardization pools turns across trajectories without requiring multiple rollouts of the same task.

\subsection{Joint Policy and Value Optimization}
\label{sec:joint-objective}

SAPO applies two forms of supervision to the shared language model. The policy objective is evaluated at every generated token, whereas the value objectives supervise one pre-action and one post-action readout per interaction turn. All advantages and value targets are computed from the rollout policy and detached before optimization.

\paragraph{Token-Level Clipped Policy Objective.}
Consider a rollout batch $\mathcal B=\{\boldsymbol{\tau}_i\}_{i=1}^{B}$. Let $\mathcal I_{\mathcal B}=\{(i,t,j):1\leq i\leq B,\,1\leq t\leq T_i,\,1\leq j\leq M_{i,t}\}$ index its valid generated response tokens, and let $N_{\mathrm{tok}}=|\mathcal I_{\mathcal B}|$. For token $a_{i,t,j}$, the causal context is $\boldsymbol{c}_{i,t,j}=[\boldsymbol{c}^{s}_{i,t};\boldsymbol{a}_{i,t,<j}]$. The standardized turn advantage is broadcast to every token in the corresponding response, $\widehat A_{i,t,j}=\operatorname{sg}(\widehat A_{i,t})$, where $\operatorname{sg}$ denotes stop-gradient. We minimize the standard PPO clipped surrogate over generated tokens:
\begin{equation}
\begin{aligned}
    \mathcal L_{\mathrm{policy}}(\theta)
    &=-\frac{1}{N_{\mathrm{tok}}}
    \sum_{(i,t,j)\in\mathcal I_{\mathcal B}}
    \min\Biggl[\frac{\pi_\theta(a_{i,t,j}\mid\boldsymbol{c}_{i,t,j})}
    {\pi_{\theta_{\mathrm{old}}}(a_{i,t,j}\mid\boldsymbol{c}_{i,t,j})}
    \widehat A_{i,t,j},\\[-2pt]
    &\operatorname{clip}\!\left(
    \frac{\pi_\theta(a_{i,t,j}\mid\boldsymbol{c}_{i,t,j})}
    {\pi_{\theta_{\mathrm{old}}}(a_{i,t,j}\mid\boldsymbol{c}_{i,t,j})},
    1-\epsilon,1+\epsilon\right)
    \widehat A_{i,t,j}\Biggr]+\beta \mathbb{D}_{\mathrm{KL}}\!\left(
\pi_{\theta}(\cdot \mid \boldsymbol{c}_{i,t,j})
\,\|\, \pi_{\mathrm{ref}}(\cdot \mid \boldsymbol{c}_{i,t,j})
\right).
\end{aligned}
    \label{eq:sapo-policy-loss}
\end{equation}
The token mask includes all generated response tokens and excludes prompt, observation, and padding tokens. $\beta$ controls the weight of the KL
penalty encouraging proximity to a reference policy $\pi_\text{ref}$. Hence, SAPO estimates credit at the turn level but applies the likelihood ratio, clipping, and gradient update at the token level.

\paragraph{Value Regression and Policy Optimization.}
Let $\mathcal T_{\mathcal B}=\{(i,t):1\leq i\leq B,\,1\leq t\leq T_i\}$ denote the valid turns and $N_{\mathrm{turn}}=|\mathcal T_{\mathcal B}|$. For $X\in\{V,Q\}$, we define,
\begin{equation}
    v^V_{i,t}=v_\theta(\boldsymbol{c}^{s}_{i,t}),
    v^Q_{i,t}=v_\theta(\boldsymbol{c}^{sa}_{i,t}),
    \bar y^X_{i,t}=\frac{\operatorname{sg}(y^X_{i,t})}{R_{\max}}.
    \label{eq:normalized-value-targets}
\end{equation}
Thus, the readouts are optimized on the scaled target space, while Equation~\ref{eq:vq-readout} restores the reward scale when constructing temporal-difference targets and advantages. Following PPO, we limit changes in the normalized value prediction relative to its rollout-time value:
\begin{equation}
    v^{X,\mathrm{clip}}_{i,t}
    =v^{X,\mathrm{old}}_{i,t}
    +\operatorname{clip}\!\left(
    v^X_{i,t}-v^{X,\mathrm{old}}_{i,t},-\epsilon_X,\epsilon_X
    \right).
    \label{eq:clipped-value-prediction}
\end{equation}
The corresponding turn-level regression loss is,
\begin{equation}
\begin{aligned}
    \mathcal L_X(\theta)
    =\frac{1}{2N_{\mathrm{turn}}}
    \sum_{(i,t)\in\mathcal T_{\mathcal B}}
    \max\Bigl(&\left(v^X_{i,t}-\bar y^X_{i,t}\right)^2,\left(v^{X,\mathrm{clip}}_{i,t}-\bar y^X_{i,t}\right)^2\Bigr),~X\in\{V,Q\}.
\end{aligned}
    \label{eq:clipped-value-loss}
\end{equation}
This clipping constrains the update in the regression objective; it does not bound the scalar readout itself. Averaging over turns also prevents response length from changing the relative weight of value supervision.

The final loss combines policy optimization, the value objectives, and entropy regularization, i.e.,
$ \mathcal{L}_{\methodname{}}(\theta)
    =\mathcal{L}_{\mathrm{policy}}
    +c_V\mathcal{L}_V
    +c_Q\mathcal{L}_Q
    -c_H\mathcal{H}(\pi_\theta)$. The policy, state-value, and action-value terms use separate masks but update the same transformer and language-model head in one backward pass. The $c_V$ and $c_Q$ control the weighting of value learning, determined empirically without further tuning, while entropy term controls exploration. Hyperparameters can be found in Appendix \ref{Implementation}.

\section{Experiments} 

We evaluate \methodname{} on two multi-turn environments and organize our experiments around four research questions: (1) \textit{As a low-cost PPO-style training paradigm, can \methodname{} match or surpass conventional PPO in task performance and training stability?} (2) \textit{How much does \methodname{} reduce the peak memory usage and runtime overhead of PPO?} (3) \textit{How does \methodname{} compare with GRPO and recent strong baselines?} (4) \textit{How much does each component contribute, practically how effectively does \methodname{} learn value estimates?}

\subsection{Experimental Setup} 
\paragraph{Benchmarks and Baselines.} We train and evaluate \methodname{} on two challenging multi-turn benchmarks, ALFWorld~\citep{shridhar2020alfworld} and WebShop~\citep{yao2022webshop}. ALFWorld evaluates long-horizon textual reasoning and decision-making in an embodied household environment, with $3{,}827$ task instances spanning six categories. WebShop evaluates interactive web navigation and product search over nearly $1.1$ million products and $12{,}000$ user instructions. We compare \methodname{} with baselines covering complementary agent and optimization paradigms. Prompting: GPT-4o~\citep{achiam2023gpt} and Gemini-2.5-Pro~\citep{team2023gemini}, together with ReAct~\citep{yao2022react} and Reflexion~\citep{shinn2023reflexion}. RL training: PPO~\citep{schulman2017proximal}, GRPO~\citep{shao2024deepseekmath}, RLOO~\citep{ahmadian2024back}, together with recent methods with dense signals: GiGPO~\citep{feng2026group}, EMPG~\citep{wang2026harnessing}, and GraphGPO~\citep{cheng2026beyond}. Further descriptions of these baselines are provided in Appendix~\ref{app:baselines}.

\paragraph{Implementation Details.} We use Qwen2.5-1.5B-Instruct, Qwen2.5-7B-Instruct, and Qwen3-14B-Instruct base models. For fair comparisons among the PPO-based and Group-based methods, we follow the matched experimental protocol of Feng et al.~\citep{feng2026group}, especially using the same number of parallel environments. Notably, \textbf{PPO employs turn-level GAE consistented with SAPO, rather than the token-level GAE reported in GiGPO} ~\citep{feng2026group}. All methods use a learning rate of $1\times10^{-6}$, a KL loss coefficient $\beta$ of $0.01$, and $150$ optimization steps. We cap each episode at $50$ turns for ALFWorld and $15$ for WebShop. We empirically set the value loss weights in SAPO to $c_V=c_Q=0.1$, without further tuning. Training is performed using NVIDIA H200, H20, and A40 GPUs. Full hyperparameters and details are provided in the Appendix \ref{Implementation}.

\begin{table*}[th]
\centering
\resizebox{\textwidth}{!}{
\begin{tabular}{llccccccc|cc}
\toprule
\multirow{2}{*}{Type} & \multirow{2}{*}{Method} & \multicolumn{7}{c|}{\textbf{ALFWorld}} & \multicolumn{2}{c}{\textbf{WebShop}} \\
 & & Pick & Look & Clean & Heat & Cool & Pick2 & All & Score & Succ.\\
\midrule
\multicolumn{10}{l}{\textit{Closed-Source Model}} \\
Prompting& GPT-4o & 75.3 & 60.8 & 31.2 & 56.7 & 21.6 & 49.8 & 48.0& 31.8 & 23.7\\
Prompting& Gemini-2.5-Pro & 92.8 & 63.3 & 62.1 & 69.0 & 26.6 & 58.7 & 60.3& 42.5 & 35.9\\
\midrule
\multicolumn{9}{l}{\textit{Qwen2.5-1.5B-Instruct}} \\
Prompting& ReAct & 17.4 & 20.5 & 15.7 & 6.2 & 7.7 & 2.0 & 12.8& 40.1& 11.3\\
Prompting& Reflexion & 35.3 & 22.2 & 21.7 & 13.6 & 19.4 & 3.7 & 21.8 & 55.8& 21.9\\

RL Training& RLOO & 88.3\textsubscript{\textpm3.0} & 52.8\textsubscript{\textpm8.6} & 71.0\textsubscript{\textpm5.9} & 62.8\textsubscript{\textpm8.7} & 66.4\textsubscript{\textpm5.5} & 56.9\textsubscript{\textpm4.7} & 69.7\textsubscript{\textpm2.5}& 73.9\textsubscript{\textpm5.6}& 52.1\textsubscript{\textpm6.7}\\

RL Training& EMPG & 85.5 & 33.5 & 78.9 & 76.2 & 74.7 & {89.1} & 73.7 & 80.4 & 60.8\\

RL Training& GiGPO\textsubscript{w/ std} & 94.4\textsubscript{\textpm5.9} & 67.5\textsubscript{\textpm4.6} & {94.8}\textsubscript{\textpm3.8} & {94.4}\textsubscript{\textpm7.8} & {79.8}\textsubscript{\textpm4.7} & 76.4\textsubscript{\textpm5.4} & {86.7}\textsubscript{\textpm1.7} & 83.1\textsubscript{\textpm1.6}& 65.0\textsubscript{\textpm3.2}\\
RL Training& GiGPO\textsubscript{w/o std} & 96.0\textsubscript{\textpm1.4} & {76.5}\textsubscript{\textpm3.9} & 91.8\textsubscript{\textpm5.5} & 91.3\textsubscript{\textpm6.3} & 71.7\textsubscript{\textpm8.4} & {79.5}\textsubscript{\textpm7.7} & 86.1\textsubscript{\textpm4.7} & {83.5}\textsubscript{\textpm1.8} & {67.4}\textsubscript{\textpm4.5}\\
RL Training& GraphGPO & 95.2\textsubscript{\textpm1.6} & \textbf{100.0}\textsubscript{\textpm0.0} & 85.3\textsubscript{\textpm2.6} & 85.7\textsubscript{\textpm5.83} & 96.3\textsubscript{\textpm2.6} & 93.7\textsubscript{\textpm2.24} & 92.7\textsubscript{\textpm1.3} & \textbf{89.3}\textsubscript{\textpm1.5} & \textbf{78.7}\textsubscript{\textpm3.9}\\
\rowcolor{gray!20}RL Training& GRPO & 85.3\textsubscript{\textpm1.5} & 53.7\textsubscript{\textpm8.0} & 84.5\textsubscript{\textpm6.8} & 78.2\textsubscript{\textpm7.9} & 59.7\textsubscript{\textpm5.0} & 53.5\textsubscript{\textpm5.6} & 72.8\textsubscript{\textpm3.6}& 75.8\textsubscript{\textpm3.5} & 56.8\textsubscript{\textpm3.8}\\
\rowcolor{gray!20}RL Training& PPO  & \textbf{100.0}\textsubscript{\textpm0.0} & {79.5}\textsubscript{\textpm4.2} & {97.9}\textsubscript{\textpm2.5} & \textbf{100.0}\textsubscript{\textpm0.0} & {81.4}\textsubscript{\textpm1.8} & \textbf{94.5}\textsubscript{\textpm4.2} & {93.0}\textsubscript{\textpm1.1}& 82.7\textsubscript{\textpm1.2} & {65.6}\textsubscript{\textpm 2.7} \\
\rowcolor{gray!20}RL Training& \textbf{\methodname{}}  & {96.4}\textsubscript{\textpm1.4} & {81.2}\textsubscript{\textpm5.1} & \textbf{100.0}\textsubscript{\textpm0.0} & \textbf{100.0}\textsubscript{\textpm0.0} & \textbf{97.2}\textsubscript{\textpm3.2} & {87.7}\textsubscript{\textpm5.1} & \textbf{94.6}\textsubscript{\textpm1.4} & 80.1\textsubscript{\textpm2.1} & {66.0}\textsubscript{\textpm 1.5} \\

\midrule
\multicolumn{9}{l}{\textit{Qwen2.5-7B-Instruct}} \\
Prompting& ReAct & 48.5 & 35.4 & 34.3 & 13.2 & 18.2 & 17.6 & 31.2 & 46.2 & 19.5\\
Prompting& Reflexion & 62.0 & 41.6 & 44.9 & 30.9 & 36.3 & 23.8 & 42.7& 58.1& 28.8\\

RL Training& RLOO & 87.6\textsubscript{\textpm4.3} & 78.2\textsubscript{\textpm8.3} & 87.3\textsubscript{\textpm5.8} & 81.3\textsubscript{\textpm7.6} & 71.9\textsubscript{\textpm5.2} & 48.9\textsubscript{\textpm8.4} & 75.5\textsubscript{\textpm4.6} & 80.3\textsubscript{\textpm3.2} & 65.7\textsubscript{\textpm4.0}\\
RL Training& EMPG & 92.9 & 75.2 & 74.8 & 86.3 & 73.7 & 65.3 & 78.5 & 81.0 & 69.3\\ 
RL Training& GiGPO\textsubscript{w/ std} & {97.7}\textsubscript{\textpm1.6} & 82.7\textsubscript{\textpm7.9} & {98.8}\textsubscript{\textpm1.6} & 83.7\textsubscript{\textpm7.2} & {89.3}\textsubscript{\textpm8.2} & 79.2\textsubscript{\textpm6.6} & {90.8}\textsubscript{\textpm1.3} & 84.4\textsubscript{\textpm2.9} & 72.8\textsubscript{\textpm3.2}\\
RL Training& GiGPO\textsubscript{w/o std} & 91.8\textsubscript{\textpm5.4} & {88.6}\textsubscript{\textpm6.3} & 95.9\textsubscript{\textpm3.2} & {90.2}\textsubscript{\textpm2.6} & 86.5\textsubscript{\textpm5.5} & {85.2}\textsubscript{\textpm7.5} & 90.2\textsubscript{\textpm2.3} & {86.2}\textsubscript{\textpm2.6} & {75.2}\textsubscript{\textpm3.8}\\
RL Training& GraphGPO & \textbf{100.0}\textsubscript{\textpm0.0} & \textbf{100.0}\textsubscript{\textpm0.0} & 91.4\textsubscript{\textpm1.5} & 92.9\textsubscript{\textpm5.8} & 94.4\textsubscript{\textpm0.0} & 92.1\textsubscript{\textpm2.2} & 95.3\textsubscript{\textpm1.1} & 86.9\textsubscript{\textpm0.7} & 80.3\textsubscript{\textpm1.3}\\
\rowcolor{gray!20}RL Training& GRPO & 90.8\textsubscript{\textpm5.1} & 66.1\textsubscript{\textpm6.7} & 89.3\textsubscript{\textpm5.4} & 74.7\textsubscript{\textpm6.9} & 72.5\textsubscript{\textpm5.4} & 64.7\textsubscript{\textpm7.3} & 77.6\textsubscript{\textpm5.2} & 79.3\textsubscript{\textpm2.8} & 66.1\textsubscript{\textpm3.7}\\ 
\rowcolor{gray!20}RL Training& PPO & \textbf{100.0}\textsubscript{\textpm0.0} & \textbf{100.0}\textsubscript{\textpm0.0} & {97.9}\textsubscript{\textpm3.6} & {88.4}\textsubscript{\textpm5.3} & {89.7}\textsubscript{\textpm5.8} & \textbf{98.7}\textsubscript{\textpm2.2} & {95.8}\textsubscript{\textpm1.2} & 87.7\textsubscript{\textpm1.8}& 82.3\textsubscript{\textpm2.2}\\
\rowcolor{gray!20}RL Training& \textbf{\methodname{}} & {98.8}\textsubscript{\textpm1.6} & {85.5}\textsubscript{\textpm11.2} & \textbf{99.1}\textsubscript{\textpm2.0} & \textbf{100.0}\textsubscript{\textpm0.0} & \textbf{95.6}\textsubscript{\textpm4.7} & {92.3}\textsubscript{\textpm8.2} & \textbf{96.1}\textsubscript{\textpm1.1} & \textbf{88.4}\textsubscript{\textpm1.7} & \textbf{83.6}\textsubscript{\textpm1.9}\\

\midrule
\multicolumn{9}{l}{\textit{Qwen3-14B-Instruct}} \\
\rowcolor{gray!20}RL Training& GRPO & {99.0}\textsubscript{\textpm1.8} & {69.1}\textsubscript{\textpm5.8} & {85.9}\textsubscript{\textpm5.0} & {79.7}\textsubscript{\textpm12.8} & {75.6}\textsubscript{\textpm24.5} & {79.7}\textsubscript{\textpm13.6} & {83.8}\textsubscript{\textpm8.6} & 85.0\textsubscript{\textpm6.1} & 77.2\textsubscript{\textpm6.6} \\ 
\rowcolor{gray!20}RL Training& PPO & \textbf{100.0}\textsubscript{\textpm0.0} & {93.6}\textsubscript{\textpm6.3} & \textbf{100.0}\textsubscript{\textpm0.0} & {86.7}\textsubscript{\textpm23.1} & \textbf{100.0}\textsubscript{\textpm0.0} & \textbf{100.0}\textsubscript{\textpm0.0} & {97.9}\textsubscript{\textpm2.3} & \textbf{92.4}\textsubscript{\textpm1.5}& \textbf{84.8}\textsubscript{\textpm1.8}\\
\rowcolor{gray!20}RL Training& \textbf{\methodname{}} & {99.0}\textsubscript{\textpm1.8} & \textbf{95.8}\textsubscript{\textpm7.2} & \textbf{100.0}\textsubscript{\textpm0.0} & \textbf{97.6}\textsubscript{\textpm4.1} & {98.7}\textsubscript{\textpm2.2} & \textbf{100.0}\textsubscript{\textpm0.0} & \textbf{98.7}\textsubscript{\textpm0.5}& {90.0}\textsubscript{\textpm1.3} & {84.4}\textsubscript{\textpm0.8}\\

\bottomrule
\end{tabular}
}
\caption{Performance results on ALFWorld and WebShop. For each RL training method, we report the mean and standard deviation over three random seeds. The ALFWorld contains six categories: Pick \& Place (Pick), Examine in Light (Look), Clean \& Place (Clean), Heat \& Place (Heat), Cool \& Place (Cool), and Pick Two \& Place (Pick2). The baseline GiGPO\textsubscript{w/o std} replaces the group std with one ~\citep{feng2026group}.}
\vspace{-10pt}
\label{tab:main}
\end{table*}

\subsection{Main Results}

\begin{figure}[t]
   \begin{center}
   \includegraphics[width=1.0\linewidth]{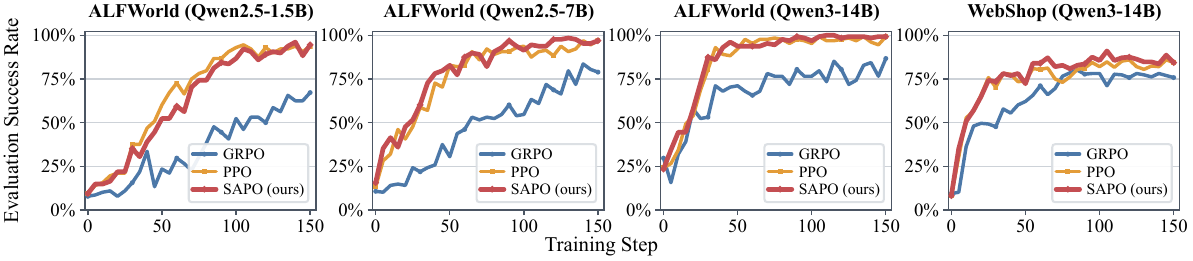}
   \end{center}
   \vspace{-10pt}
\caption{\textbf{Evaluation Comparison.} We show the evaluation performance curves among SAPO, PPO and GRPO throughout training under several representative experimental settings.}
   \label{fig:eval_main}
   \vspace{-15pt}
\end{figure}


\textbf{ALFWorld and WebShop Results.}~~Table~\ref{tab:main} and Figure \ref{fig:eval_main} show that \textbf{SAPO achieves PPO-level performance across both benchmarks and all model scales while significantly outperforming GRPO in terms of the best performance, convergence speed, and learning stability}. On ALFWorld, SAPO reaches aggregate success rates of $94.6\%$, $96.1\%$, and $98.7\%$ at 1.5B, 7B, and 14B, slightly exceeding PPO at each scale and surpassing GRPO by $14.9$--$21.8$ points. On WebShop, SAPO remains within $1.3$ points of PPO in success rate but improves over GRPO by $7.2$--$17.5$ points. Moreover, as illustrated in Figure~\ref{fig:eval_main}, the performance curves for SAPO closely align with those of PPO, with SAPO showing a slight advantage in certain more complex tasks (notably in WebShop). Additionally, both methods exhibit significantly lower learning variance compared to GRPO, a trend even more pronounced in the training curves (see Figure \ref{fig:train_main} in Appendix \ref{appendix_train_main}). Furthermore, compared to the baselines with dense credit signals, SAPO achieves the best aggregate ALFWorld result at 1.5B, although GraphGPO remains stronger on WebShop. At 7B, SAPO leads all dense-signal baselines, reaching $96.1\%$ ALFWorld success, an $88.4$ WebShop score, and $83.6\%$ WebShop success. 


\begin{figure}[h]
   \begin{center}
   \includegraphics[width=1.0\linewidth]{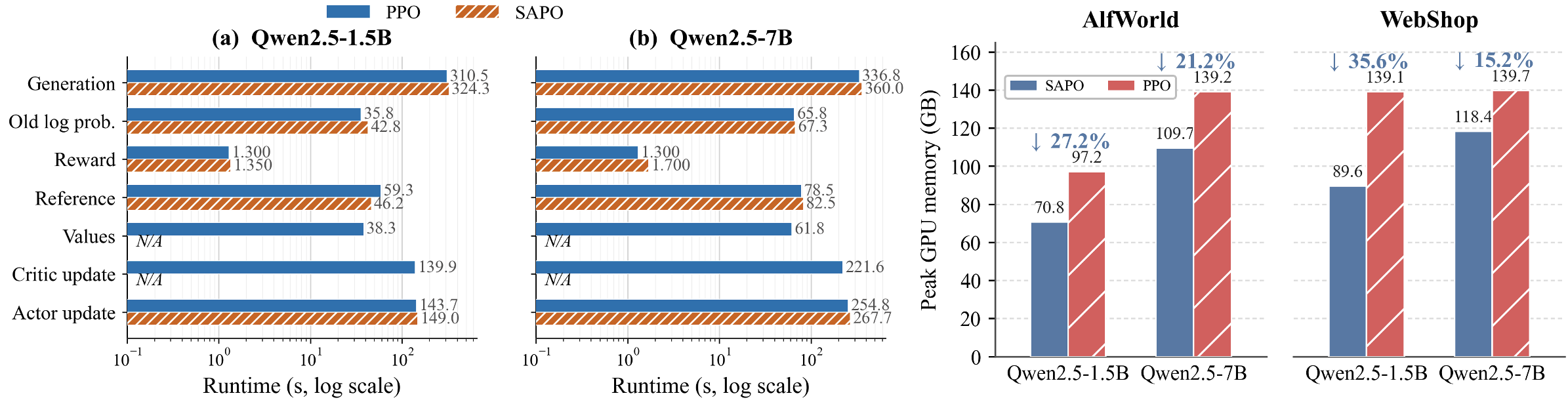}
   \end{center}
   \vspace{-15pt}
\caption{\textbf{Computational resource comparison between SAPO and PPO} using Qwen2.5-1.5/7B on ALFWorld and WebShop under fully matched experimental settings, e.g., batch size and sampling budgets. The left panel reports the per-iteration runtime breakdown across training stages in seconds on a logarithmic scale, where N/A denotes stages not used by SAPO. The right panel reports peak GPU memory usage in GB during training on NVIDIA H200 GPUs; the 1.5B experiments use one GPU and the 7B experiments use two GPUs. Each experiment group used the same server.}
   \label{fig:peak_gpu}
   \vspace{-5pt}
\end{figure}

\textbf{Computational Resource Comparison.}~~Figure~\ref{fig:peak_gpu} compares the training cost of SAPO and PPO under matched settings. SAPO consistently lowers peak GPU memory across both environments and model scales, with reductions ranging from $15.2\%$ to $35.6\%$. The largest reduction occurs on WebShop with Qwen2.5-1.5B, where peak memory drops from $139.1$\,GB to $89.6$\,GB; the reduction persists at 7B and on ALFWorld. These savings arise from SAPO's unified actor--critic design, which performs value learning within the policy model rather than maintaining a separate critic and its optimizer state.

The runtime comparison on ALFWorld reflects both this architectural simplification and SAPO's autoregressive rollout design. Relative to PPO, SAPO reduces the measured per-iteration time by $22.6\%$ at 1.5B and $23.6\%$ at 7B. Beyond removing direct critic computation, each autoregressive rollout supplies both policy-gradient and value-learning signals through causal-boundary readouts, allowing the two objectives to be optimized jointly in one shared-model update. SAPO therefore avoids repeated sequence processing through a second model and the independent critic optimization that together consume roughly one quarter of a PPO iteration at both scales. The shared update makes SAPO's actor-update stage only modestly slower, leaving a substantial net reduction in runtime.

In summary, by synthesizing performance results and resource comparisons, we provide empirical evidence demonstrating the high efficiency and low computational cost of SAPO, as well as its stable scalability across base models ranging from 1.5B to 14B parameters. \textbf{These findings demonstrate that SAPO effectively balances policy performance with resource overhead, validating its viability as a cost-effective alternative to conventional actor–critic PPO frameworks across the evaluated tasks}.


\begin{table}[ht]
\centering
\vspace{-15pt}
\caption{Ablation study comparing variants without action value learning loss ($-\cal {L}_{Q}$), advantage normalization, the scaled value setting and with a single logit. The best performance under the same setting is highlighted in bold. We report the average success rate (\%) from three seeds.}
\label{ablation}
\resizebox{1.0\textwidth}{!}{
\setlength{\tabcolsep}{2.5mm}{
\begin{tabular}{l|ccccccc}
\toprule
\multirow{2}{*}{\textbf{Method}}
& \multicolumn{7}{c}{\textbf{ALFWorld}} \\
& Pick & Clean & Cool & Look & Heat & Pick2 & All \\
\midrule
 \methodname{} (1.5B)  & {96.4}\textsubscript{\textpm1.4} & {81.2}\textsubscript{\textpm5.1} & \textbf{100.0}\textsubscript{\textpm0.0} & \textbf{100.0}\textsubscript{\textpm0.0} & \textbf{97.2}\textsubscript{\textpm3.2} & \textbf{87.7}\textsubscript{\textpm5.1} & \textbf{94.6}\textsubscript{\textpm1.4}   \\
$-\cal {L}_{Q}$ (1.5B) &  \textbf{100.0}\textsubscript{\textpm0.0} & {81.6}\textsubscript{\textpm13.3} & {98.6}\textsubscript{\textpm2.4} & \textbf{100.0}\textsubscript{\textpm0.0} & {85.3}\textsubscript{\textpm9.4} & {84.3}\textsubscript{\textpm5.6} & {92.2}\textsubscript{\textpm0.8}\\
$-$ Norm (1.5B) &  {97.1}\textsubscript{\textpm2.7} & \textbf{84.1}\textsubscript{\textpm14.1} & {91.7}\textsubscript{\textpm10.3} & {96.7}\textsubscript{\textpm4.5} & {84.8}\textsubscript{\textpm7.4} & {83.6}\textsubscript{\textpm8.6} & {89.1}\textsubscript{\textpm3.4}  \\
$-$ Value~Scaling (1.5B) &  {95.3}\textsubscript{\textpm3.2} & {70.6}\textsubscript{\textpm25.5} & {88.3}\textsubscript{\textpm9.5} & {88.3}\textsubscript{\textpm11.5} & {82.9}\textsubscript{\textpm1.7} & {45.2}\textsubscript{\textpm6.8} & {80.5}\textsubscript{\textpm2.9}  \\
$w/$~Single Logit (1.5B) &  {99.0}\textsubscript{\textpm1.9} & {83.4}\textsubscript{\textpm17.3} & {95.5}\textsubscript{\textpm6.4} & \textbf{100.0}\textsubscript{\textpm0.0} & {85.7}\textsubscript{\textpm6.1} & {73.9}\textsubscript{\textpm10.0} & {89.9}\textsubscript{\textpm3.3}  \\

\midrule
 \methodname{} (7B)  & {98.8}\textsubscript{\textpm1.6} & {85.5}\textsubscript{\textpm11.2} & {99.1}\textsubscript{\textpm2.0} & \textbf{100.0}\textsubscript{\textpm0.0} & \textbf{95.6}\textsubscript{\textpm4.7} & \textbf{92.3}\textsubscript{\textpm8.2} & \textbf{96.1}\textsubscript{\textpm1.1}   \\
$-\cal {L}_{Q}$ (7B)&  {95.9}\textsubscript{\textpm4.4} & {82.7}\textsubscript{\textpm4.3} & {96.2}\textsubscript{\textpm6.6} & {97.4}\textsubscript{\textpm4.4} & {82.3}\textsubscript{\textpm5.4} & {87.4}\textsubscript{\textpm11.0} & {91.2}\textsubscript{\textpm2.0}  \\
$-$ Norm (7B) &  \textbf{100.0}\textsubscript{\textpm0.0} & \textbf{94.7}\textsubscript{\textpm7.3} & \textbf{100.0}\textsubscript{\textpm0.0} & {98.3}\textsubscript{\textpm3.3} & {91.4}\textsubscript{\textpm3.5} & {89.0}\textsubscript{\textpm2.0} & {95.5}\textsubscript{\textpm2.0}  \\
$-$ Value~Scaling (7B) &  {89.9}\textsubscript{\textpm4.1} & {71.7}\textsubscript{\textpm7.1} & {88.4}\textsubscript{\textpm8.4} & {91.3}\textsubscript{\textpm6.6} & {90.4}\textsubscript{\textpm7.7} & {72.6}\textsubscript{\textpm15.7} & {85.7}\textsubscript{\textpm2.1}  \\
$w/$~Single Logit (7B) &  \textbf{100.0}\textsubscript{\textpm0.0} & {86.9}\textsubscript{\textpm9.4} & \textbf{100.0}\textsubscript{\textpm0.0} & {93.1}\textsubscript{\textpm7.0} & {83.5}\textsubscript{\textpm7.3} & {83.9}\textsubscript{\textpm5.0} & {92.2}\textsubscript{\textpm1.7}  \\

\bottomrule
\end{tabular}
}
}
\vspace{-10pt}
\end{table}

\subsection{Ablation Experiments and Value Analysis}

\textbf{Ablation Study.}~~~Table~\ref{ablation} ablates four SAPO components on ALFWorld. Removing value-target scaling causes the largest aggregate success drop at both 1.5B and 7B and increases variation, underscoring its importance for stable policy--value co-training in a shared language model. This pattern is consistent with scaled targets preventing large value-logit shifts that could interfere with policy learning. Removing the auxiliary $\mathcal{L}_Q$ or replacing the two-logit readout with a single logit also degrades performance at both scales, supporting action-value supervision and the dual-logit representation. Advantage normalization has a stronger effect at 1.5B than at 7B, indicating that normalizing advantages across the rollout batch can also improve policy performance.

\begin{wrapfigure}{r}{0.6\columnwidth}
\vspace{-10pt}
  \centering
  \includegraphics[width=\linewidth]{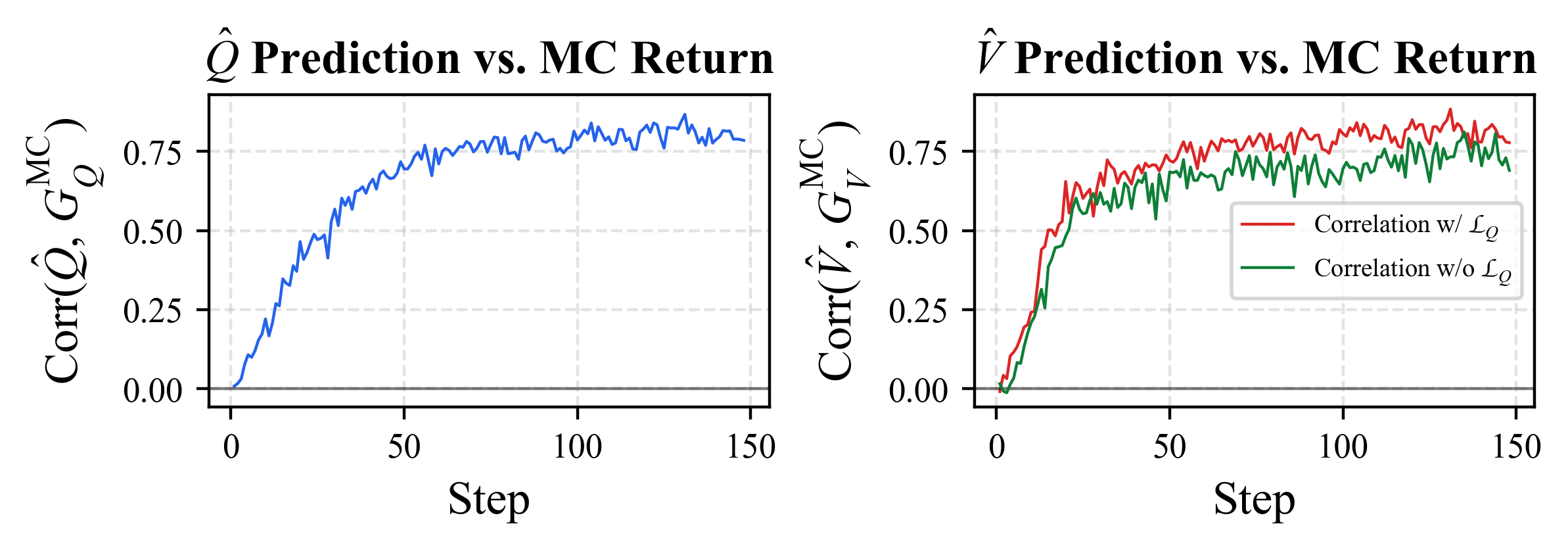}
\vspace{-15pt}
  \caption{\textbf{Pearson correlations} of predicted $\hat Q$ value (left) and predicted $\hat V$ value (right) with their corresponding Monte Carlo return estimates.}
\vspace{-7pt}
  \label{fig:qv_mc_cor}
\end{wrapfigure}

\textbf{Learned Value Correlations.}~~Figure~\ref{fig:qv_mc_cor} tracks the Pearson correlation between learned value predictions and their corresponding Monte Carlo returns of interactive trajectories throughout training. The correlations for both the predicted action value $\hat Q~(v^Q)$ and the predicted state value $\hat V~(v^V)$ rise steadily, suggesting that both become better at estimating expected discounted cumulative rewards. The right panel further compares $\hat V$ correlations with and without auxiliary ${\cal L}_Q$ learning. We observe that auxiliary ${\cal L}_Q$ learning yields a higher correlation for the $\hat V$ curve (red), potentially improving value learning. These results suggest that, despite sharing a single language model, the two causal-boundary readouts can learn return-aligned value estimates and serve the role of conventional value models. They also provide empirical support for the designed auxiliary ${\cal L}_Q$ learning as a means of improving $\hat V$ learning. Together, these findings support SAPO’s central motivation, i.e., stable value learning comparable to that of separate critic models while sharing the LLM weights.

\section{Conclusion}
In summary, we presented SAPO, a single-rollout autoregressive framework that unifies policy learning and value learning within a single causal language model. By leveraging joint policy, value, and on-policy SARSA objectives, SAPO preserves rigorous temporal credit assignment while entirely eliminating the need for a separate critic and multiple rollouts. Our extensive evaluation demonstrates that SAPO consistently matches standard PPO and outperforms GRPO in efficiency, scalability, and training stability, significantly reducing both memory and time overhead. Ultimately, SAPO establishes a highly practical, low-cost post-training paradigm for long-horizon LLM agents.

\subsection*{AI use statement}
We used generative AI tools to assist with language editing, paraphrasing, and improving the clarity and readability of the manuscript. All AI-assisted revisions were reviewed and edited by the authors to ensure that they accurately reflected the intended technical meaning and did not introduce unsupported claims. The authors take full responsibility for the final content of this work, including all text, claims, and artifacts produced with the assistance of generative AI.

\subsection*{Reproducibility statement}
To support reproducibility, we describe SAPO and its optimization procedure in the main text. The experimental setup and appendix document the base models, training configurations, reward design, and computational resources used in our experiments. The appendix also provides the prompt templates for ALFWorld and WebShop, specifying the information available to the agent and the required response format. Together, these descriptions provide the implementation and evaluation details needed to reproduce our results.

\bibliography{iclr2027_conference}
\bibliographystyle{iclr2027_conference}


\appendix
\section{SAPO Training Algorithm}
\label{app:sapo-algorithm}

Algorithm~\ref{alg:sapo} presents the complete SAPO training procedure. The rollout parameters remain fixed while trajectories and learning targets are constructed. Policy and value predictions are then recomputed together by the current model and optimized through one shared update.

\begin{algorithm}[H]
\caption{Single-Rollout Autoregressive Policy Optimization}
\label{alg:sapo}
\begin{algorithmic}[1]
\STATE \textbf{Input} trainable model $f_\theta$, reference policy $\pi_{\mathrm{ref}}$, task distribution $\mathcal{D}$, and optimization hyperparameters
\FOR{training iteration $k=1,\ldots,K$}
    \STATE Set $\theta_{\mathrm{old}}\leftarrow\theta$ and sample a task batch $\mathcal{B}\sim\mathcal{D}$
    \FOR{each task $\boldsymbol{x}\in\mathcal{B}$}
        \STATE Sample one trajectory $\boldsymbol{\tau}$ with $\pi_{\theta_{\mathrm{old}}}$ over the action vocabulary $\mathcal{W}_{\mathrm{act}}$
        \FOR{each turn $t=1,\ldots,T$ in $\boldsymbol{\tau}$}
            \STATE Store the old token log probabilities, reward $r_t$, and termination indicator $d_t$
            \STATE Read $V_t^{\mathrm{old}}$ before the action and $Q_t^{\mathrm{old}}$ after the action from the shared value basis
        \ENDFOR
        \STATE Set $V_{T+1}^{\mathrm{old}}=Q_{T+1}^{\mathrm{old}}=A_{T+1}^{\mathrm{GAE}}=0$
        \FOR{$t=T,\ldots,1$}
            \STATE Compute TD error: $\delta_t\leftarrow r_t+\gamma(1-d_t)V_{t+1}^{\mathrm{old}}-V_t^{\mathrm{old}}$
            \STATE Compute GAE: $A_t^{\mathrm{GAE}}\leftarrow\delta_t+\gamma\lambda(1-d_t)A_{t+1}^{\mathrm{GAE}}$
            \STATE Compute $V$ target: $y_t^V\leftarrow V_t^{\mathrm{old}}+A_t^{\mathrm{GAE}}$
            \STATE Compute $Q$ target: $y_t^Q\leftarrow r_t+\gamma(1-d_t)Q_{t+1}^{\mathrm{old}}$
        \ENDFOR
    \ENDFOR
    \STATE Apply the invalid action adjustment when available and normalize advantages over all valid turns in $\mathcal{B}$
    \STATE Broadcast each normalized turn advantage $\widehat{A}_t$ to the valid tokens of its action
    \STATE Run one shared-model forward pass to obtain current policy log probabilities, $V_\theta$, and $Q_\theta$
    \STATE Compute $\mathcal{L}_{\mathrm{policy}}$, $\mathcal{L}_V$, and $\mathcal{L}_Q$, scaling the value targets by $R_{\max}$
    \STATE Compute reference-policy regularization $\mathcal{L}_{\mathrm{KL}}$ and policy entropy $\mathcal{H}$
    \STATE Update $\theta$ by minimizing $\mathcal{L}_{\mathrm{policy}}+c_V\mathcal{L}_V+c_Q\mathcal{L}_Q-c_H\mathcal{H}$
\ENDFOR
\STATE \textbf{Output} optimized parameters $\theta$
\end{algorithmic}
\end{algorithm}

\section{Training Curves}\label{appendix_train_main}
\begin{figure}[h]
   \begin{center}
   \includegraphics[width=1.0\linewidth]{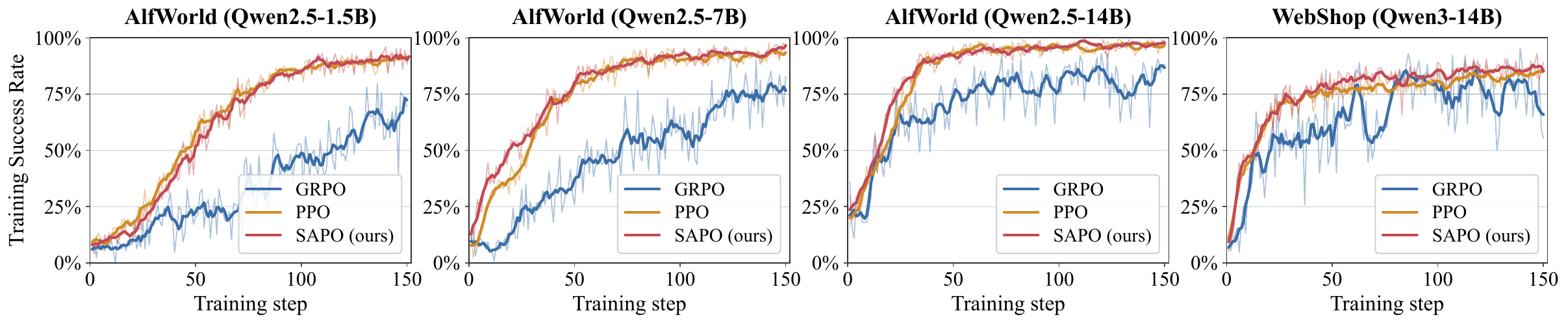}
   \end{center}
   \vspace{-10pt}
\caption{\textbf{Learning Curve Comparison.} We show the learning performance curves among SAPO, PPO and GRPO throughout training under several representative experimental settings.}
   \label{fig:train_main}
   \vspace{-10pt}
\end{figure}

Figure~\ref{fig:train_main} shows that SAPO closely tracks PPO across all four settings. On ALFWorld, both methods improve rapidly and reach similar success rates at 1.5/7/14B, whereas GRPO generally learns more slowly and remains below them through most of training. WebShop shows a similar pattern: SAPO and PPO reach comparable levels, while GRPO exhibits larger fluctuations, particularly in later steps. Furthermore, in most tasks, especially the more complex WebShop tasks, SAPO slightly outperforms PPO. Taken together, these curves suggest that SAPO preserves PPO-like optimization behavior across tasks and model scales despite sharing policy and value learning within a single model.

\section{Implementation Details}\label{Implementation}
\subsection{Training Details and Hyperparameters}
We implement SAPO using verl-agent (GiGPO), a scalable framework built on VeRL for training LLM agents through multi-turn interactions. All experiments run on x86\_64 systems with Ubuntu 20.04/22.04. Our compute infrastructure includes 6 NVIDIA H200 GPUs and two NVIDIA H20 clusters with 64 and 8 GPUs, respectively. We use Qwen2.5-1.5B-Instruct, Qwen2.5-7B-Instruct, and Qwen3-14B-Instruct as base models, with checkpoint sizes of approximately 3.1, 15.2, and 29.6 GB, respectively. We train Qwen3-14B-Instruct on 32 H20 GPUs, Qwen2.5-7B-Instruct on 2 H200 GPUs, and Qwen2.5-1.5B-Instruct on either 4 A40 GPUs or a single H200 GPU.

The specific settings for SAPO are as follows. The total training loss comprises the value loss and Q-loss, with weights of $c_V=0.1$ and $c_Q=0.1$, respectively, without further hyperparameter tuning. A temperature coefficient is used to adjust the sensitivity of the predicted value; a value of 1.0 was adopted for both ALFWorld and WebShop, again without further tuning. The \texttt{value\_cliprange} for the predicted value $v(c)$ is set to 0.2, consistent with the standard \texttt{cliprange} used in PPO.

\paragraph{ALFWorld.}
We train on ALFWorld for 150 iterations with a batch size of 256. Each episode is capped at 50 interaction turns, with maximum response and prompt lengths of 512 and 2048 tokens, respectively. The rollout sampling temperature is set to 0.4 during training and 1.0 during validation. PO-style SAPO and PPO use the same number of parallel environments as group-based baselines to ensure a comparable training samples. For group-based methods, each iteration samples 16 groups of 8 rollouts, using $16 \times 8 = 128$ parallel environments. For SAPO and PPO, each iteration samples identical 128 rollouts, using $128$ parallel environments. We set both lambda $\lambda$ and gamma $\gamma$ in the generalized advantage estimation to 0.95 and the KL-divergence loss coefficient to $0.01$. The entropy coefficient $c_H$ is set in accordance with the existing GiGPO method. The maximum trajectory-level outcome reward is $R_{\max}=10$, and each invalid action incurs a turn-level penalty of $-0.1$.

\paragraph{WebShop.}
We train on WebShop for 150 iterations with a batch size of 64 and a maximum of 15 interaction turns per episode. The response length is capped at 512 tokens, while the prompt length is capped at 4096 tokens to accommodate the longer webpage observations. We set both lambda $\lambda$ and gamma $\gamma$ in the generalized advantage estimation to 0.95, and we use the same KL-divergence loss coefficient ($0.01$), training and validation sampling temperatures, parallel rollout configuration, entropy coefficient and reward design as in ALFWorld.

\subsection{Environmental Prompts} 

\begin{figure}[h]
\centering
\resizebox{0.999\textwidth}{!}{
\begin{tcolorbox}[colback=gray!5!white, colframe=gray!75!gray, 
title=Prompt Template for ALFWorld, boxrule=0.3mm, width=\textwidth, arc=3mm, auto outer arc=true]
You are an expert agent operating in the ALFRED embodied Environment. Your task is to: \textcolor{brown}{\{task\_description\}}. Prior to this step, you have already taken \textcolor{brown}{\{step\_count\}} step(s). Below are the most recent \textcolor{brown}{\{history\_length\}} observations and the corresponding actions you took: \textcolor{brown}{\{action\_history\}}. You are now at step \textcolor{brown}{\{current\_step\}} and your current observation is: \textcolor{brown}{\{current\_observation\}}. Your admissible actions of the current situation are: [\textcolor{brown}{\{admissible\_actions\}}].

Now it's your turn to take an action.
You should first reason step-by-step about the current situation. This reasoning process MUST be enclosed within \textcolor{deepgreen}{<think>} \textcolor{deepgreen}{</think>} tags.
Once you've finished your reasoning, you should choose an admissible action for current step and present it within \textcolor{deeppurple}{<action>} \textcolor{deeppurple}{</action>} tags.
\end{tcolorbox}
}
\caption{The prompt template for ALFWorld environment.}
\label{prompt:alfworld_prompt_temp}
\vspace{-10pt}
\end{figure}

\begin{figure}[h]
\centering
\resizebox{0.999\textwidth}{!}{
\begin{tcolorbox}[colback=gray!5!white, colframe=gray!75!gray, 
title=Prompt Template for WebShop, boxrule=0.3mm, width=\textwidth, arc=3mm, auto outer arc=true]
You are an expert autonomous agent operating in the WebShop e-commerce environment. Your task is to: \textcolor{brown}{\{task\_description\}}. Prior to this step, you have already taken \textcolor{brown}{\{step\_count\}} step(s). Below are the most recent \textcolor{brown}{\{history\_length\}} observations and the corresponding actions you took: \textcolor{brown}{\{action\_history\}}. You are now at step \textcolor{brown}{\{current\_step\}} and your current observation is: \textcolor{brown}{\{current\_observation\}}. Your admissible actions for the current situation are: [\textcolor{brown}{\{available\_actions\}}].
Now it's your turn to take one action for the current step.
You should first reason step-by-step about the current situation, then think carefully which admissible action best advances the shopping goal. This reasoning process MUST be enclosed within \textcolor{deepgreen}{<think>} \textcolor{deepgreen}{</think>} tags.
Once you've finished your reasoning, you should choose an admissible action for current step and present it within \textcolor{deeppurple}{<action>} \textcolor{deeppurple}{</action>} tags.
\end{tcolorbox}
}
\caption{The prompt template for WebShop environment.}
\label{prompt:webshop_prompt_temp}
\vspace{-10pt}
\end{figure}

\paragraph{Prompt Templates for ALFWorld and WebShop.}
Figure~\ref{prompt:alfworld_prompt_temp} shows the prompt template for ALFWorld. At each turn, the prompt includes the task description, the number of completed steps, a recent observation--action history, the current observation, and the admissible actions. This context helps the agent track changes in the environment and choose an action appropriate to the current state. The agent is instructed to enclose its step-by-step reasoning in \texttt{<think>...</think>} and return exactly one admissible action in \texttt{<action>...</action>}, separating the reasoning text from the command executed by the environment.

The WebShop template, shown in Figure~\ref{prompt:webshop_prompt_temp}, follows the same structure, providing the shopping objective, the current step index, a recent observation--action history, the current webpage observation, and the available actions. The history contextualizes previous searches, navigation steps, and product selections, while the available actions identify the valid operations at the current step. Using the same output format, the agent explains its choice in \texttt{<think>...</think>} and places a single action that advances the shopping objective in \texttt{<action>...</action>}.

\section{Baseline Descriptions}
\label{app:baselines}

We compare SAPO with prompting-based agents, standard policy-gradient algorithms, and recent methods designed for fine-grained credit assignment. The first group measures what can be achieved through inference-time reasoning alone, while the latter two evaluate SAPO against established and specialized RL training paradigms.

\paragraph{Prompting-based Agents.}
\textbf{GPT-4o}~\citep{achiam2023gpt} and \textbf{Gemini-2.5-Pro}~\citep{team2023gemini} are closed-source general-purpose models evaluated through prompting without task-specific RL training. They provide reference points for strong proprietary models operating without task-specific RL training. \textbf{ReAct}~\citep{yao2022react} interleaves natural-language reasoning with environment actions, allowing the model to update its plan after each observation. \textbf{Reflexion}~\citep{shinn2023reflexion} augments this interaction loop with verbal self-reflection: after an unsuccessful attempt, the model summarizes its errors and conditions subsequent attempts on this feedback. ReAct and Reflexion therefore improve behavior through prompting and test-time interaction rather than parameter updates.

\paragraph{Standard Policy-Gradient Baselines.}
\textbf{PPO}~\citep{schulman2017proximal} is our principal actor--critic baseline. It trains a policy with the clipped surrogate objective and uses a separately learned value function with GAE to propagate delayed rewards across interaction turns. PPO provides a strong reference for explicit value learning and temporal credit assignment, but incurs the memory and computation required by an additional critic. \textbf{GRPO}~\citep{shao2024deepseekmath} removes the learned critic by sampling a group of trajectories for each task and standardizing their outcome rewards within the group. The resulting relative advantage is used to optimize the policy, reducing critic-related training cost at the expense of multiple rollouts and trajectory-level credit. \textbf{RLOO}~\citep{ahmadian2024back} is another critic-free REINFORCE-style method. For each sampled trajectory, it uses the mean reward of the remaining trajectories as a leave-one-out baseline, reducing gradient variance without fitting a value model.

\paragraph{Fine-Grained Credit-Assignment Baselines.}
\textbf{GiGPO}~\citep{feng2026group} combines an episode-level macro advantage with a step-level micro advantage obtained by grouping actions that originate from matched anchor states across trajectories. We report both its standard variant and the variant that omits standard-deviation normalization, denoted GiGPO\textsubscript{w/ std} and GiGPO\textsubscript{w/o std}, respectively. \textbf{EMPG}~\citep{wang2026harnessing} modulates policy gradients using step-wise predictive entropy and the final task outcome. It strengthens updates for confident correct actions, penalizes confident errors, and attenuates highly uncertain steps. \textbf{GraphGPO}~\citep{cheng2026beyond} aggregates sampled trajectories into a state-transition graph and estimates each visited state's distance to a successful goal. These methods provide direct comparisons with approaches that derive denser step-level supervision from trajectory groups, predictive uncertainty, or graph structure, whereas SAPO obtains turn-level learning signals from a learned value function using a single rollout per task.

\section{AlfWorld Cases}

\begin{figure}[th]
       \vspace{-20pt}
   \centering
   \makebox[\linewidth][c]{%
      \includegraphics[width=0.92\linewidth]{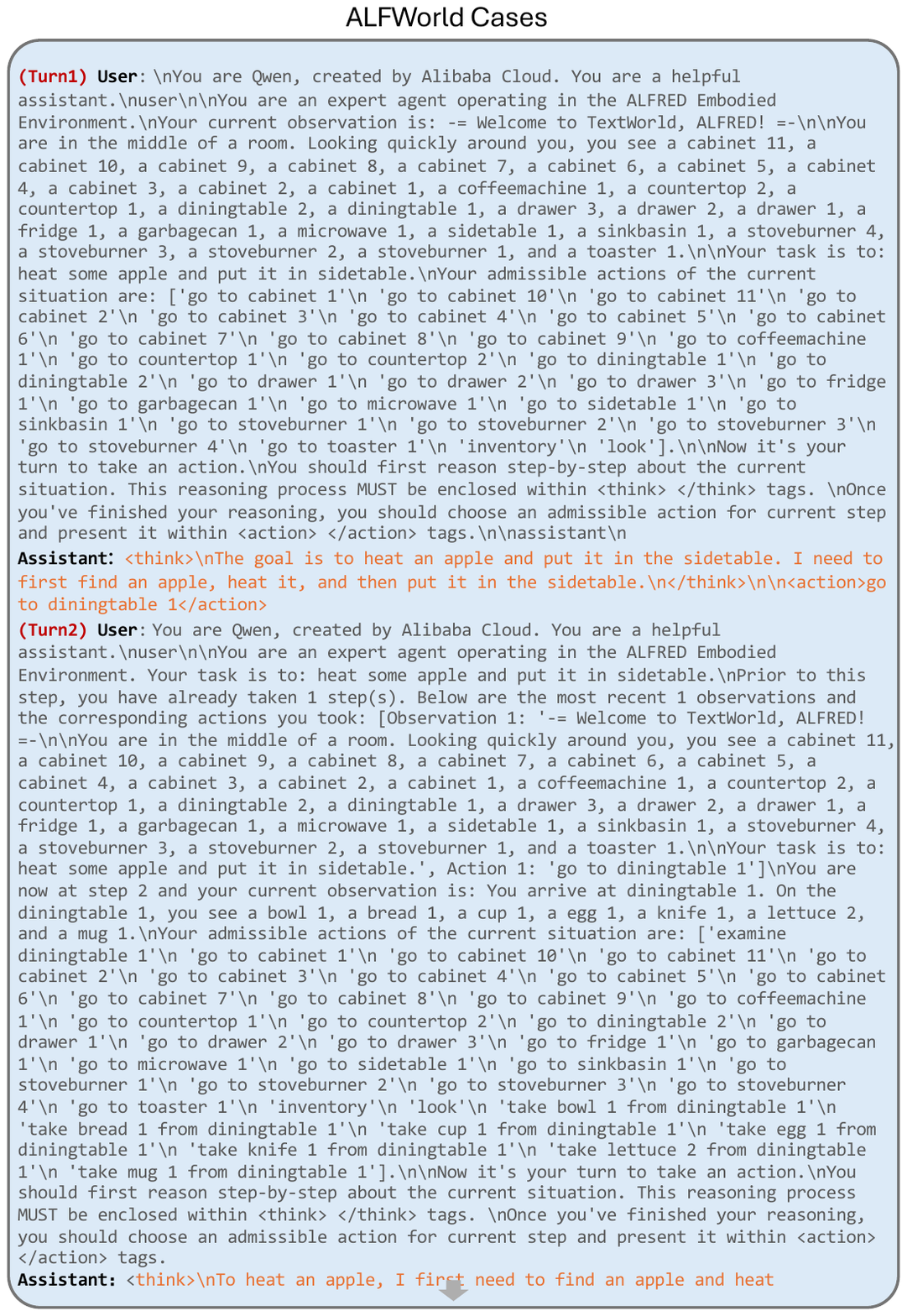}%
   }
   \vspace{-50pt}
   \caption{Case study of SAPO on ALFWorld using Qwen2.5-1.5B-Instruct.}
   \label{fig:alf9}
\end{figure}
\begin{figure}[th]
   \centering
   \makebox[\linewidth][c]{%
      \includegraphics[width=1.2\linewidth]{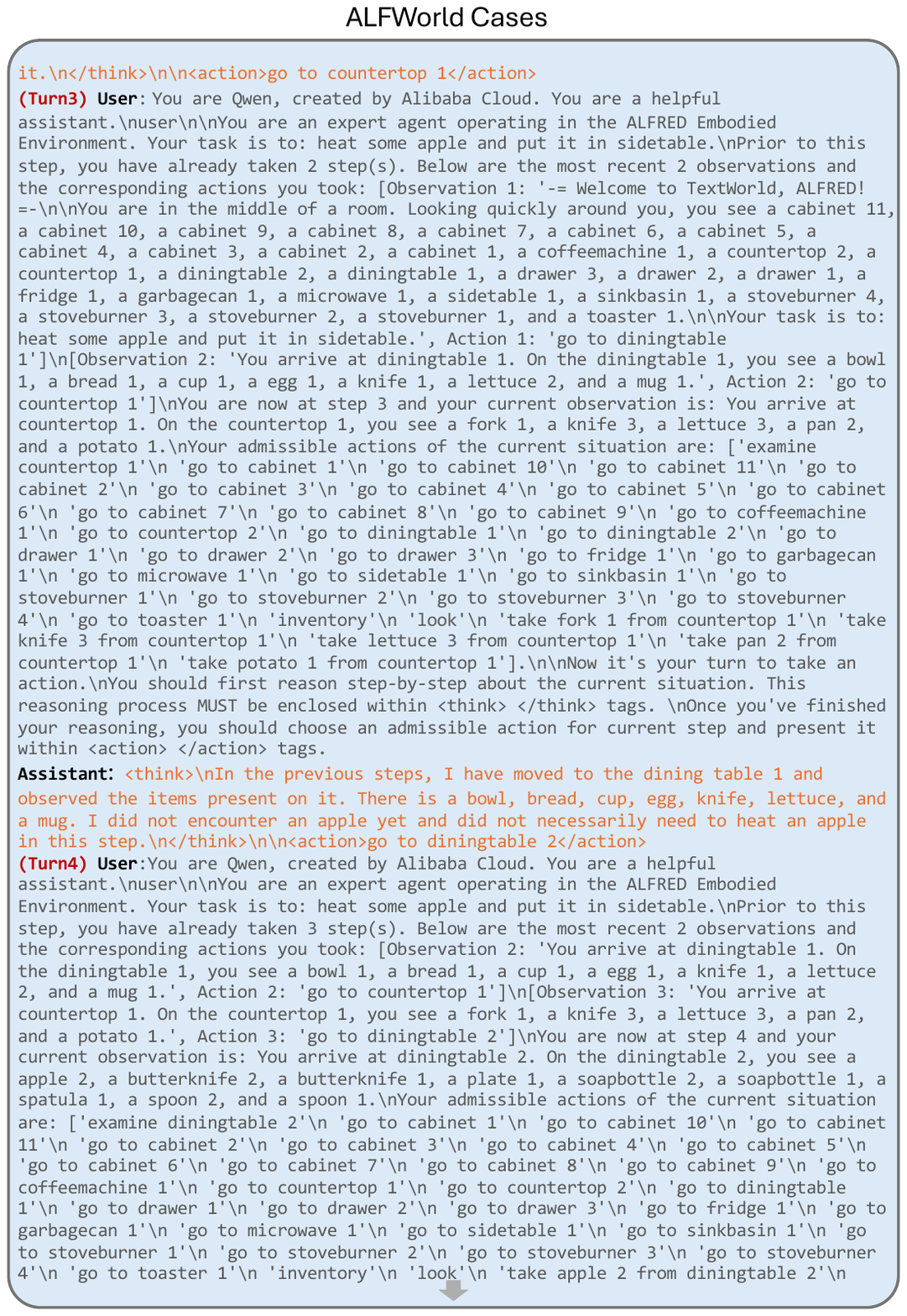}%
   }
   \vspace{-55pt}
   \caption{Case study of SAPO on ALFWorld using Qwen2.5-1.5B-Instruct.}
   \label{fig:alf10}
\end{figure}\begin{figure}[th]
   \centering
   \makebox[\linewidth][c]{%
      \includegraphics[width=1.2\linewidth]{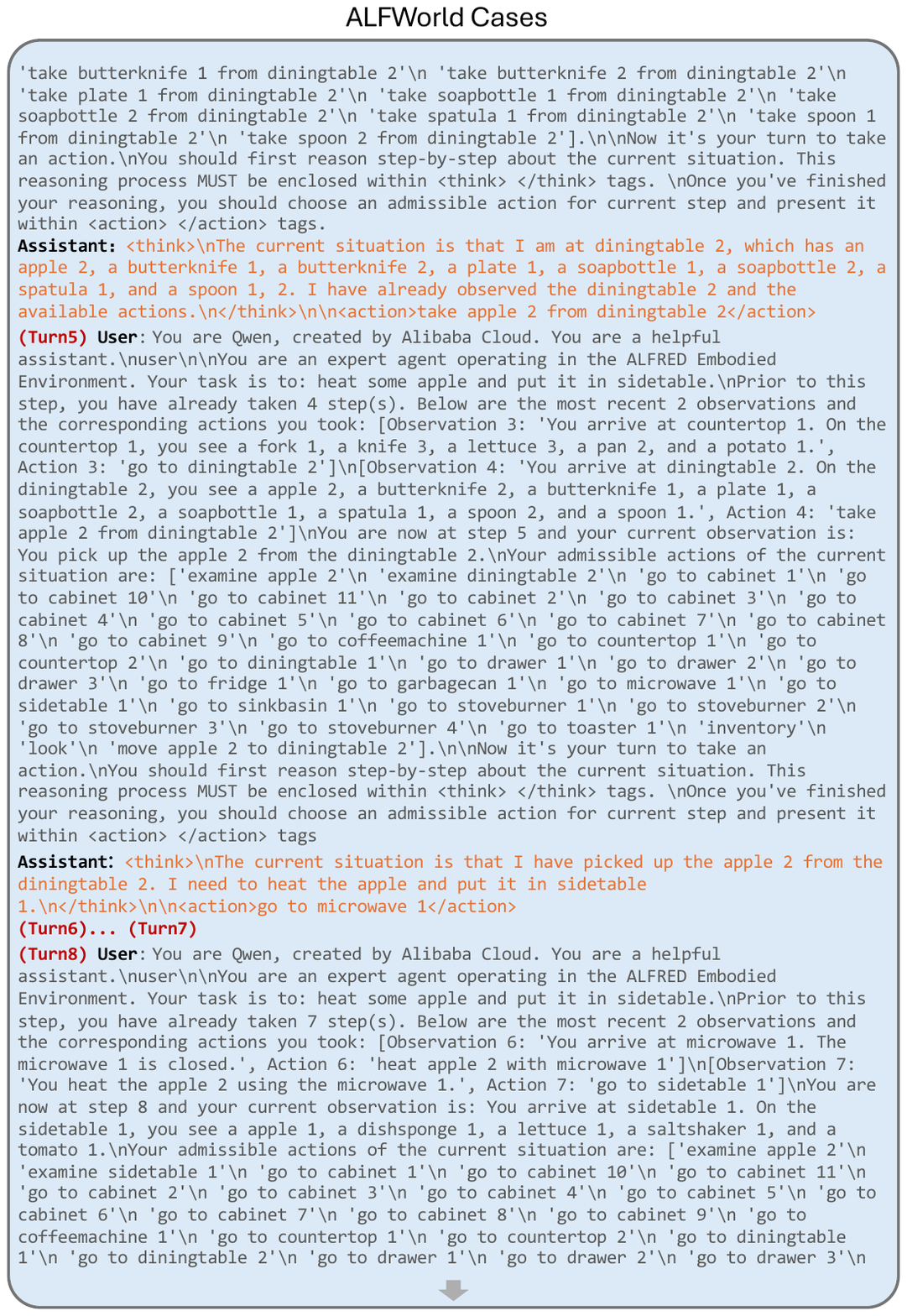}%
   }
   \vspace{-55pt}
   \caption{Case study of SAPO on ALFWorld using Qwen2.5-1.5B-Instruct.}
   \label{fig:alf11}
\end{figure}

\begin{figure}[th]
   \centering
   \makebox[\linewidth][c]{%
      \includegraphics[width=1.2\linewidth]{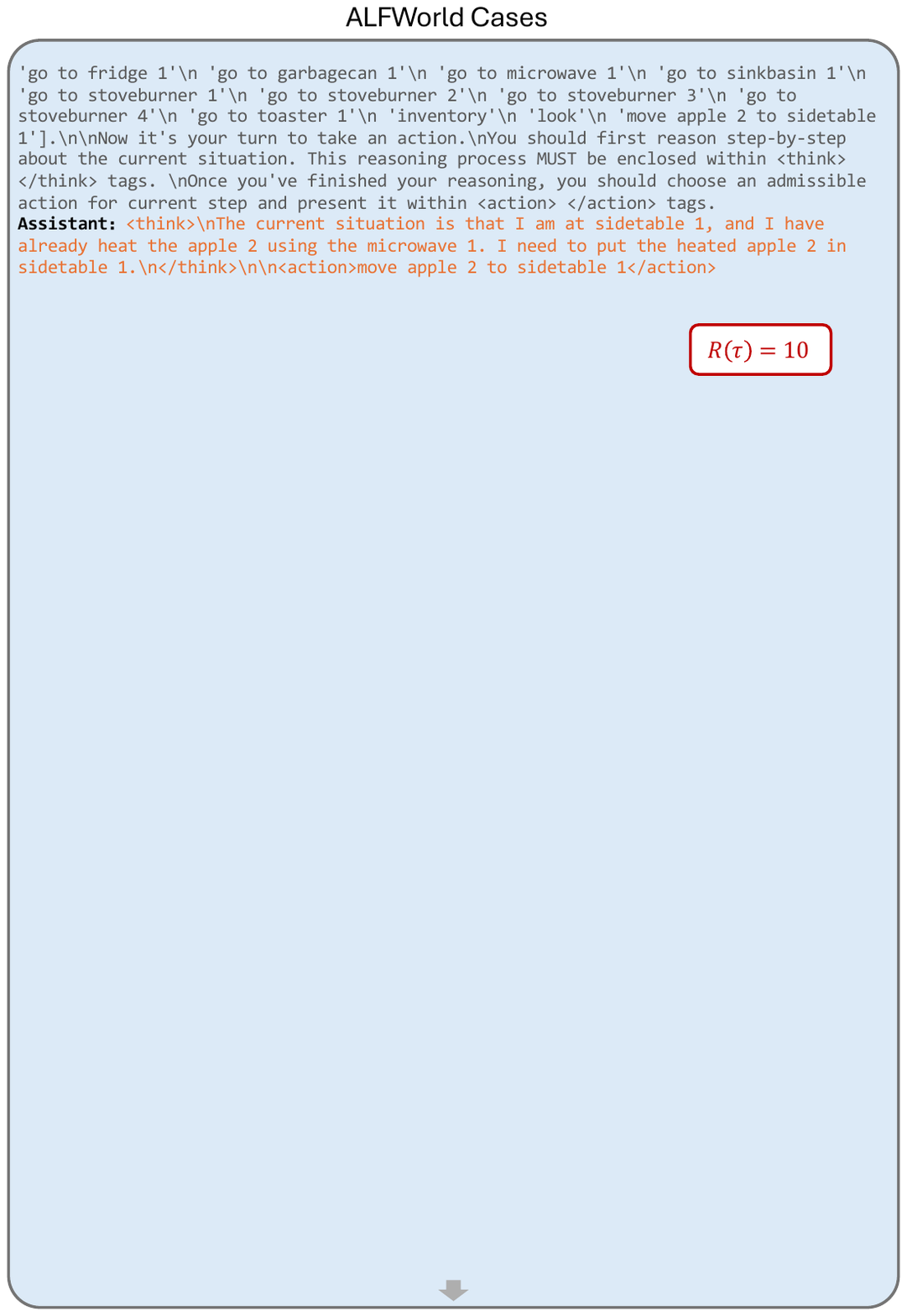}%
   }
   \vspace{-55pt}
   \caption{Case study of SAPO on ALFWorld using Qwen2.5-1.5B-Instruct.}
   \label{fig:alf12}
\end{figure}

\end{document}